\documentclass[12pt]{article}

\usepackage{CJKutf8}

\usepackage{scicite}

\usepackage{times}
\usepackage[utf8]{inputenc} 
\usepackage{hyperref}       
\usepackage{url}            
\usepackage{booktabs}       
\usepackage{amsfonts}       
\usepackage{nicefrac}       
\usepackage{microtype}      
\usepackage{multicol}
\usepackage{color}
\usepackage{pgfplots}
\usepackage{graphicx}
\usepackage{makecell} 
\usepackage{multirow}
\usepackage{booktabs, graphicx, amssymb}
\usepackage{longtable}
\usepackage{array}
\usepackage{stfloats}
\usepackage{balance}
\usepackage{multicol}
\usepackage{color}
\usepackage{epstopdf}
\usepackage{bm}
\usepackage{amsmath}
\usepackage{amssymb}
\usepackage{booktabs} 
\usepackage{epsfig}
\usepackage{enumitem}
\usepackage{cleveref}
\usepackage{arydshln}
\usepackage{balance}
\usepackage{xspace}
\usepackage{wrapfig}
\usepackage{enumitem}
\pgfplotsset{compat=newest}
\usepackage{subfigure}

\newcommand{\hide}[1]{} 

\usepackage[subfigure]{tocloft}

\cftsetindents{figure}{0em}{4.5em}
\cftsetindents{table}{0em}{4em}

\newenvironment{sciabstract}{%
\begin{quote} \bf}
{\end{quote}}

\title{Efficiently Building a Domain-Specific Large Language Model from Scratch: A Case Study of a Classical Chinese Large Language Model}

\author{
Shen Li\thanks{\href{mailto:shen@mail.bnu.edu.cn}{shen@mail.bnu.edu.cn}}, Renfen Hu, Lijun Wang\\
\normalsize{Beijing Normal University}
}
\date{}

\begin{document} 
\begin{CJK*}{UTF8}{bsmi}

\baselineskip 24pt

\maketitle 

\begin{sciabstract}

General-purpose large language models demonstrate notable capabilities in language comprehension and generation, achieving results that are comparable to, or even surpass, human performance in many natural language processing tasks. Nevertheless, when general models are applied to some specific domains, e.g., Classical Chinese texts, their effectiveness is often unsatisfactory, and fine-tuning open-source foundational models similarly struggles to adequately incorporate domain-specific knowledge. To address this challenge, this study developed a large language model, AI Taiyan, specifically designed for understanding and generating Classical Chinese. Experiments show that with a reasonable model design, data processing, foundational training, and fine-tuning, satisfactory results can be achieved with only 1.8 billion parameters. In key tasks related to language processing of Classical Chinese such as punctuation, identification of allusions, explanation of word meanings, and translation between ancient and modern Chinese, this model exhibits a clear advantage over both general-purpose large models and domain-specific traditional models, achieving levels close to or surpassing human baselines. This research provides a reference for the efficient construction of specialized domain-specific large language models. Furthermore, the paper discusses the application of this model in fields such as the collation of ancient texts, dictionary editing, and language research, combined with case studies.
\end{sciabstract}

\section{Introduction}
\label{sec:intro} 

In recent years, general-purpose large language models (LLMs), represented by GPT-4 \cite{achiam2023gpt} and LLAMA2 \cite{touvron2023llama}, have attracted widespread attention from both the academic community and the general public \cite{zhao2023survey}. However, these general-purpose LLMs are primarily designed and developed for general users, with their training corpora largely sourced from publicly available internet data, which contain relatively limited domain-specific knowledge. For example, Chinese data on the internet mainly comes from news websites, forums, and similar sources, and is almost exclusively expressed in modern Chinese. This inevitably restricts the ability of models to understand and process Classical Chinese.

As a carrier of Chinese cultural heritage, Classical Chinese texts embody a wealth of linguistic and cultural knowledge. Given that literary expression in Classical Chinese has spanned thousands of years, its characters, vocabulary, grammar, and phonological systems have continuously evolved, with each era producing complex linguistic phenomena that pose significant challenges to comprehension of modern readers. In addition to linguistic difficulties, understanding Classical Chinese texts also relies on extra historical and cultural knowledge. For instance, ancient authors often used multiple variants to refer to people and events, favored implicit and nuanced expression, and frequently drew upon allusions to convey emotions or ideas. The dual challenges of language and culture not only complicate the learning and reading of Classical Chinese for contemporary readers but also constitute major obstacles in the collation of ancient texts and research in natural language processing of Classical Chinese.

To better support the collation of ancient texts, the teaching of Classical Chinese, and digital humanities research, we have developed a large language model specifically designed for Classical Chinese, named ``AI Taiyan'' (AI 太炎) \footnote{\href{https://t.shenshen.wiki}{https://t.shenshen.wiki}}. This paper introduces the construction of this model, covering its architectural design, data processing, base training, and fine-tuning. We further explore its potential applications in areas such as ancient text collation, lexicography, and linguistic research through case studies.
\section{Related Work}
\label{sec:rel}

\subsection{Natural Language Processing in Classical Chinese}

In recent years, driven by the practical needs of ancient text collation and publication, Classical Chinese teaching, and digital humanities research, studies on natural language processing of Classical Chinese have become increasingly in-depth. These studies encompass a wide range of tasks, including punctuation, word segmentation, part-of-speech tagging, named entity recognition, relation extraction, word sense disambiguation, and translation between Classical and Modern Chinese.

Since ancient texts generally do not use punctuation marks, one of the key tasks in the collation of ancient texts is sentence segmentation and punctuation. According to \cite{hu2021}, most existing ancient text data have not yet been punctuated. For example, in the Dazhige Ancient Literature Collection 2.0 corpus \footnote{\href{https://guji.chat}{https://guji.chat}}, which contains approximately 3.3 billion characters, only about 25\% of the data include punctuation, highlighting the enormous scale of ancient text collation and the urgent need for automatic punctuation technologies. To address this issue, researchers have achieved promising results using traditional machine learning models \cite{zhang2009}, neural network models \cite{wang2017}, and pre-trained language models \cite{yu2019,hu2021,yuan2022}.

For large-scale lexical-level research on ancient texts using computational techniques, fundamental Chinese information processing tasks such as Classical Chinese word segmentation, part-of-speech tagging, named entity recognition, and word sense disambiguation are often involved. The development of Classical Chinese word segmentation has gone through three stages: rule-based methods \cite{qiu2008}, statistical methods \cite{liang2013}, and machine learning and deep learning-based methods \cite{huang2015,cheng2020,tang2022slepen}. Part-of-speech tagging and named entity recognition follow similar research paradigms. For word sense analysis, \cite{shu2021} constructed a Classical Chinese word sense annotation corpus with more than one million characters and developed a word sense disambiguation algorithm based on a pre-trained Classical Chinese language model.

Translation between Classical and Modern Chinese (also known as Wenyan–Baihua translation) is a comprehensive task that integrates many of the challenges in understanding Classical Chinese, and it is in high demand to support modern readers in reading and learning Classical texts. \cite{liu2019ancient} developed a Classical-to-Modern translation system based on the Transformer model. \cite{guo2023towards} further proposed a translation model leveraging disyllabic word alignment and a dual-masked Transformer model, achieving better performance than the model by \cite{liu2019ancient}.

In summary, neural network models and pre-trained language model methods have been widely applied in the field of Classical Chinese processing. However, these methods impose high requirements on the scale and quality of training data. In Classical Chinese processing scenarios, many tasks are characterized by being ``low-resource'' but ``knowledge-intensive'', posing significant challenges for language resource development and model research. Existing studies have shown that mechanisms such as data augmentation and joint learning can effectively alleviate these issues \cite{MESS202104002}. Given the strong capabilities of large language models in multi-task and few-shot learning, building dedicated large language models for Classical Chinese holds great potential for enhancing the comprehensive capabilities of Classical Chinese.

\subsection{Construction of Large Language Models for Specialized Domains}

To enhance large language models’ understanding of domain-specific knowledge, methods such as retrieval-augmented generation (RAG) and the construction of domain-specific models have emerged. Retrieval-augmented generation does not require modification of model parameters; instead, it primarily utilizes vector retrieval and matching techniques to obtain relevant domain knowledge from external knowledge bases or databases. The retrieved information is then combined with the current query and input into the language model, enabling it to answer questions with reference to external knowledge \cite{gao2023retrieval}. In comparison, building large language models tailored to specific domains allows for more systematic learning of domain knowledge, thereby supporting vertical applications, e.g., ChatLaw for the legal domain \cite{cui2023chatlaw}, medGPT \footnote{\href{https://medgpt.co}{https://medgpt.co}} for the medical domain, and ``Mozi'' \footnote{\href{https://github.com/gmftbyGMFTBY/science-llm}{https://github.com/gmftbyGMFTBY/science-llm}} for scientific literature. In the field of Classical Chinese, models such as ``Xunzi'' \footnote{\href{https://github.com/Xunzi-LLM-of-Chinese-classics/XunziALLM}{https://github.com/Xunzi-LLM-of-Chinese-classics/XunziALLM}} and ``Jiusi'' have also been developed, aiming to facilitate the analysis and processing of ancient texts. However, most of the aforementioned domain-specific large language models are obtained by continued training or fine-tuning of open-source general-domain models such as LLaMA, Qwen, and Baichuan. \cite{taylor2022galactica} and \cite{lehman2023we} have pointed out that models trained on specialized domain data tend to perform better on domain-specific tasks. Therefore, this study aims to construct a large language model specialized in Classical Chinese, ``AI Taiyan'', which is designed to more comprehensively encode knowledge of Classical Chinese language and culture.
\section{Methods}
\label{sec: method}

\subsection{Architecture}
Drawing on the latest advancements in large language model architectures, our model is based on the Transformer framework, utilizing the SwiGLU activation function \cite{shazeer2020glu} and adopting ALiBi positional encoding \cite{press2021train} to better handle long text sequences. To accelerate the training process, we incorporate the Flash Attention mechanism \cite{dao2022flashattention}.

With respect to model size, mainstream open-source large language models typically adopt configurations of 6–7B, 13–14B, or 70B parameters. Considering that our Classical Chinese language model is designed to encode domain-specific knowledge, the scale of training data is substantially smaller than that used for general English or modern Chinese tasks. Therefore, the model size should be aligned with the data scale. Following the optimal model structure scaling curve proposed by \cite{hoffmann2022training}, we design our model with 52 layers (blocks) and a total of 1.8 billion (1.8B) parameters. Recent studies have demonstrated that, with appropriate architectural design and training, smaller large language models can achieve a favorable balance between efficiency and performance, as exemplified by Gemma (2B) \footnote{\href{https://ai.google.dev/gemma}{https://ai.google.dev/gemma}} and MiniCPM (2.4B) \footnote{\href{https://github.com/OpenBMB/MiniCPM}{https://github.com/OpenBMB/MiniCPM}}.

\subsection{Pretraining}
The training of large language models typically involves two stages. The first is the pre-training stage, in which the base language model is trained using large-scale unlabeled text data to acquire fundamental language abilities. The second is the supervised fine-tuning stage, where the model is guided to learn domain-specific knowledge and perform a range of specific tasks using a substantial amount of labeled data.

During pre-training, the model primarily acquires foundational linguistic knowledge through the task of next-word prediction. As a machine learning model, the parameter weights of a large language model are determined by fitting the training data; thus, the scale, quality, and diversity of the training data are critical determinants of the model’s language capabilities. Although large-scale models for Classical Chinese focus on the understanding and generation of ancient Chinese texts, their outputs are often intended for contemporary readers. For example, tasks such as Classical-to-Modern translation require the conversion of Classical Chinese texts into modern Chinese, so the base language model must be proficient in both ancient and modern Chinese expressions. Additionally, both simplified and traditional Chinese texts are needed for practical applications in Classical Chinese processing.

To this end, during the pre-training phase of ``AI Taiyan'', we collected approximately 225 billion characters of high-quality modern Chinese text and 3.5 billion characters of Classical Chinese text as pre-training corpora, with both simplified and traditional Chinese texts represented in certain proportions. The modern Chinese corpus includes data from online news, encyclopedias, forums, and other sources \footnote{Following these project: \href{https://github.com/Embedding/Chinese-Word-Vectors}{https://github.com/Embedding/Chinese-Word-Vectors} and \href{https://data.baai.ac.cn/datadetail/BAAI-CCI2}{https://data.baai.ac.cn/datadetail/BAAI-CCI2}.}, while the Classical Chinese corpus covers a wide range of genres such as poetry, prose, and novels \footnote{Based on this corpus: \href{https://github.com/garychowcmu/daizhigev20}{https://github.com/garychowcmu/daizhigev20}.}.

\subsection{Supervised finetuning}
During the supervised fine-tuning stage, our aim is for “AI Taiyan” to focus on domain-specific knowledge and address specialized problems within the field. Therefore, it is unnecessary for the model to be specifically trained on general conversational tasks or open-domain question answering, which are already strengths of large general-purpose language models. Instead, the emphasis should be placed on the specific challenges of understanding Classical Chinese. Accordingly, the design of supervised fine-tuning tasks follows two main principles:

First, the selected tasks should cover different levels of linguistic knowledge, including character, word, sentence, and paragraph, and should guide the model to learn classical cultural knowledge found in ancient texts. Second, for each task, it should be possible to collect high-quality and large-scale annotated data.

Based on these principles, we designed four fine-tuning tasks: segmentation and punctuation, allusion recognition, word explanation, and translation between Classical and Modern Chinese. Specific examples are shown in Table \ref{table: mission examples}.

\begin{table}
\small
\centering
\begin{tabular}{|c|c|c|}
\hline
Mission & Input & Output \\ \hline
\makecell{Segmentation \\and \\Punctuation} & \makecell{州城西南隅有黃鶴樓者\\給上述文本添加標點。\\(Add punctuation to the above text.)} & 州城西南隅，有黃鶴樓者。 \\ \hline
\makecell{Allusion\\Recognition} & \makecell{劉楨徒有氣，管輅獨無年。\\識別文本中的典故。\\(Identify the allusions in the text.)} & \makecell{劉楨徒有氣，使用了典故\\劉楨病，喻指臥病窮醜。\\(``劉楨徒有氣'' uses the allusion \\``劉楨病'' (Liu Zhen’s illness) \\refering to being bedridden and \\suffering in poverty and ugliness.)} \\ \hline
\makecell{Word\\Explanation} & \makecell{州城西南隅，有黃鶴樓者。\\文本中的“隅”是什麼意思？\\(What does ``隅'' mean in the text?)} & \makecell{角落。\\(Corner.)} \\ \hline
Translation & \makecell{州城西南隅，有黃鶴樓者。\\將上文翻譯成白話文。\\(Translate the above text \\into Modern Chinese.)} & \makecell{在鄂州城的西南角，\\有一座樓叫黃鶴樓。\\(At the southwest corner of \\Ezhou city, there is a tower \\called Yellow Crane Tower.)} \\ \hline
\end{tabular}
\caption{Examples of supervised finetuning.}
\label{table: mission examples}
\end{table}

Through data collection and augmentation, we obtained approximately 18 billion characters of training data for supervised fine-tuning, including about 1.5 billion characters for punctuation, 10 billion characters for word explanation, 200 million characters for allusion recognition, and 5.2 billion characters for Classical-to-Modern translation.

\subsection{Training Hyper-Parameters}
\cite{tang2024rethinking} pointed out that repeated utilization of data can enhance the training effectiveness of language models. Therefore, we performed a certain degree of resampling during the training of ``AI Taiyan''. According to the scaling laws proposed by \cite{hoffmann2022training}, we trained the 1.8B-parameter model on 250 billion characters of pre-training data, followed by supervised fine-tuning on an additional 50 billion characters. The maximum learning rates were set to 2e-4 for pre-training and 5e-5 for finetuning, and subsequently both decayed to 0 with a cosine schedule \cite{loshchilov2016sgdr}. To avoid gradient problems, the training is in BF16 precision.
\section{Model Performance}
\label{sec: perf}
This paper conducts open evaluations on four key tasks in Classical Chinese information processing. To ensure fairness and objectivity in the evaluation results, all evaluation datasets are sourced from the Chinese Classics Database (aka 中華經典古籍庫) \footnote{This database contains high-quality, critically edited and published ancient texts, and is equipped with rigorous anti-crawling and anti-copying mechanisms. Therefore, it is highly suitable for use as model testing data. \href{https://publish.ancientbooks.cn/docShuju/platform.jspx}{https://publish.ancientbooks.cn/docShuju/platform.jspx}} and other non-open-source internet repositories, thereby guaranteeing that the models have not encountered the test data during either pre-training or finetuning stages. In addition to evaluating ``AI Taiyan'', we introduce several existing models and the answers of graduate students majoring in literature and history as comparative baselines for each task.

For general large language models, we selected GPT-4 \footnote{The results processed by GPT-4 were obtained via the OpenAI API, using the model version gpt-4-1106-preview. In the experiments, the temperature parameter was set to 0 to ensure consistency and stability of the model's output.}, which has demonstrated outstanding performance across a range of general evaluations, and tested it on all four tasks. In terms of domain-specific models, we employed ``Xunzi'' (Xunzi-Qwen-7B-CHAT), a Classical Chinese large language model finetuned from a general open-source model, and evaluated it on punctuation and Classical-to-Modern Chinese translation tasks according to its documentation. Additionally, for the Classical-to-Modern Chinese translation task, we included Baidu Translate ``Chinese (Classical Chinese) – Chinese (Simplified)'' machine translation system as a comparative reference (traditional machine translation) \footnote{We obtained Baidu Translate' results via API. \href{https://api.fanyi.baidu.com/doc/21}{https://api.fanyi.baidu.com/doc/21}}. Finally, we invited multiple graduate students in literature and history to participate in manual evaluations for allusion recognition, word explanation, and Classical-to-Modern Chinese translation tasks. The following section introduces the specific evaluation methods and results for each task. \footnote{All contrast experiments were conducted in March, 2024. However, the results of ``AI Taiyan'' will continue to be updated once a new version is done.}

\subsection{Segmentation and Punctuation}
For the segmentation and punctuation task, we randomly selected 200 passages of carefully collated classical Chinese texts from the Chinese Classics Database as test data and evaluated the performance of each model using the F1 score \footnote{F1 score in the punctuation task is computed by exactly matching these character ``，。！？；：、'', while types are ignored in the segmentation task.}, as shown in Table 2. It is worth noting that a prominent issue with existing large language models is their inability to accurately reproduce the original text when adding punctuation marks, frequently resulting in character substitutions, omissions, or insertions. This problem persists regardless of prompt engineering. In the output results for ``Xunzi'', 20.5\% of the samples generated by the models contained errors in the original text, and samples generated by GPT-4 had an 11\% error rate in reproducing the original. In comparison, ``AI Taiyan'' was optimized for the punctuation restoration task during its decoding process by constraining the model output to only include the original vocabulary and punctuation marks, thus completely avoiding errors in reproducing the original text.

\begin{table}
\centering
\begin{tabular}{|c|c|c|c|}
\hline
Model & Text Error & Segmentation F1 & Punctuation F1 \\ \hline
Xunzi & 20.5\% & 0.9262 & 0.7644 \\ \hline
GPT-4 & 11\% & 0.8749 & 0.7250 \\ \hline
AI Taiyan & 0\% & 0.9699 & 0.8651 \\ \hline
\end{tabular}
\caption{Results of the segmentation and punctuation task. Text error means the difference between the original text and the output text of models.}
\label{table: segmentation and punctuation results}
\end{table}

To more accurately evaluate the punctuation performance, we excluded samples with output errors when calculating the punctuation results for ``Xunzi'' and GPT-4, focusing only on samples where the original text was correctly reproduced. \footnote{The F1 scores for sentence segmentation and punctuation of ``Xunzi'' and GPT-4 are calculated based only on the samples with correct outputs, that is, samples with erroneous output texts are excluded. In practice, however, the impact of the text error rate must also be taken into account, meaning that their actual F1 scores for sentence segmentation and punctuation would be even lower in real-world applications.} As shown in Table \ref{table: segmentation and punctuation results}, ``AI Taiyan'' demonstrates a clear advantage in both sentence segmentation and punctuation tasks, with the F1 score for sentence segmentation approaching 0.97, reaching a level suitable for practical application.

\subsection{Allusion Recognition}

\begin{table}
\centering
\begin{tabular}{|c|c|c|}
\hline
Model & Detection Accuracy & Identification F1 \\ \hline
Human Baseline & 81.57\% & 0.7751 \\ \hline
GPT-4 & 72.56\% & 0.0947 \\ \hline
GPT-4 + RAG & - & 0.47 \\ \hline
AI Taiyan & 89.39\% & 0.7594 \\ \hline
\end{tabular}
\caption{Results of the allusion recognition task. Detection task is to predict whether the text uses allusions or not. Identification task needs to predict which allusions are present in the text.}
\label{table: allusion results}
\end{table}

This study utilizes the dataset and evaluation methodology constructed by \cite{MESS202411003} to evaluated various models on the tasks of allusion detection and specific allusion identification. Allusion detection is formulated as a binary classification task, which determines whether a given text contains an allusion. Accuracy is used as the evaluation metric. Specific allusion identification is treated as a multi-label and multi-class classification task, which aims to identify which allusions are present in the given text. F1 score is adopted as the evaluation metric. Among the baselines, the ``Human Baseline'' refers to the average score of professional annotators on the test set \footnote{The professional annotators included graduate students majoring in Classical Chinese and senior undergraduate students majoring in Chinese Language and Literature. The test set was annotated by multiple annotators simultaneously. By comparing the annotation results of individual annotators with the final annotations included in the test set, individual annotation scores were obtained. The average of these scores across annotators was calculated to determine the human baseline. Given the considerable difficulty of allusion recognition, annotators were permitted to consult knowledge bases or reference books during the annotation process.}. ``+RAG'' denotes the introduction of a retrieval-augmented generation mechanism based on an external allusion knowledge base. As shown in Table \ref{table: allusion results} \footnote{All results come from \cite{MESS202411003} except ``AI Taiyan''.}, allusion identification is a highly challenging task that not only involves semantic understanding of the text, but also requires substantial cultural knowledge. Even annotators with a background in Chinese language studies are unable to achieve high accuracy. GPT-4, as a general-purpose large language model, performs poorly on this task, achieving an F1 score of less than 0.1 for specific allusion identification. While the introduction of retrieval augmentation with an external knowledge base leads to a significant improvement, the F1 score only reaches 0.47. In contrast, ``AI Taiyan'' surpasses professional annotators in allusion detection accuracy and achieves performance in specific allusion identification that approaches the human baseline.

\subsection{Word Explanation}
Considering the practical needs of word sense disambiguation for assisting in the collation of ancient texts and the teaching of Classical Chinese, we constructed the test set for this task from two sources: (1) We selected several of the most recently published editions of ancient texts from the Chinese Classics Database \footnote{Although we did not use the Chinese Classics Database to train our model, we made every effort to select the most recently published classical Chinese texts for the test set in order to rigorously avoid any overlap between the test data and the data seen during model training.}, and randomly extracted 100 annotation entries; (2) We selected 100 annotation entries from texts related to extracurricular reading and examinations at the secondary school level \footnote{\href{http://wyw.5156edu.com}{http://wyw.5156edu.com}}. In total, the test set consists of 200 entries. Below are two examples, with the target words to be explained marked by 【　】.
\begin{itemize}
    \item{若鉛山諸邑所造柬紙，則全用細竹料厚質蕩成，以【射】重價。最上者曰官柬，富貴之家，通刺用之，其紙敦厚而無筋膜。
    
    (If the card paper is made in places like Qianshan, it is entirely produced using fine bamboo material, resulting in thick and sturdy paper, in order to [ask for] a high price. The best kind is called ``official card paper'', which wealthy families use for sending name cards. This type of paper is thick, solid, and free of visible fibers.)}
    \item{其汞海、草汞之說，無端狂妄，【耳食】者信之。若水銀已升朱，則不可復還為汞，所謂造化之巧已盡也。
    
    (Their claims about ``Mercury Sea'' and ``Herbal Mercury'' are completely unfounded and absurd, yet some [gullible people] believe these rumors. If mercury has already been refined into cinnabar, it can no longer be turned back into mercury. This shows that the transformations of nature have reached their limit and can go no further.)}
\end{itemize}

In our experiments, in addition to ``AI Taiyan'', we also introduced GPT-4 and answers from master's and doctoral students majoring in Chinese literature and history as comparative baselines. During answering, the graduate students were not allowed to consult any reference materials and they relied solely on their contextual understanding and personal linguistic knowledge to provide explanation. Given the diversity of possible paraphrasing expressions, it is not feasible to directly calculate accuracy through character matching. Moreover, the reference answers from annotated editions of ancient texts and extracurricular readings may not be entirely precise. Therefore, we invited two graduate students majoring in Classical Chinese to manually evaluate both model-generated and human-generated answers.

To ensure fairness and reliability in the evaluation process, for each annotation, we provided three sets of anonymized and randomly ordered answers, such that the evaluators could not identify which answer was produced by which model or individual. Evaluators were also supplied with reference answers and permitted to consult various resources. Each answer was scored according to the following criteria: 1 point for correct and precise answers that facilitate understanding; 0.5 points for answers that are close but contain some issues or ambiguities; and 0 points for incorrect or misleading answers. We first conducted pilot evaluations and discussions to ensure a shared understanding of the scoring criteria among the evaluators, then proceeded to the formal evaluation. Experimental results showed that the overall scoring consistency between the two evaluators, as measured by the Spearman correlation coefficient, reached 0.8842.

\begin{table}
\centering
\begin{tabular}{|c|c|c|}
\hline
Model & Accuracy & Strict Accuracy \\ \hline
Human Baseline & 45.38\% & 36.5\% \\ \hline
GPT-4 & 54.5\% & 47\% \\ \hline
AI Taiyan & 86.5\% & 83.5\% \\ \hline
\end{tabular}
\caption{Results of the word explanation task.}
\label{table: explanation results}
\end{table}

The test results for word explanation task are shown in Table \ref{table: explanation results}. ``Strict accuracy'' refers to the proportion of completely correct answers (scored 1 point), while ``accuracy'' denotes the combined proportion of completely correct and partially correct answers (scored 0.5 points). As indicated in the table, the word explanation task remains challenging even for master's and doctoral students majoring in literature and history. However, ``AI Taiyan'' not only significantly outperforms GPT-4 and the human baseline, but also achieves an accuracy rate exceeding 80\%. This suggests that the model can be leveraged to make preliminary judgments on the meanings of difficult or important words in texts, thereby assisting users in reading or collating ancient texts. Furthermore, the related technology also holds potential for supporting dictionary compilation and research on the meanings of words in Classical Chinese.

\subsection{Classical-to-Modern Translation}
Classical-to-Modern translation is a comprehensive task that not only requires the correct interpretation of the meaning of characters, words, sentences, and paragraphs in Classical Chinese texts, but also necessitates the integration of relevant cultural background knowledge to render the text meaning in coherent and fluent modern Chinese. Given the complexity of Classical-to-Modern translation, in addition to employing conventional automatic machine translation evaluation methods, we also incorporated human evaluation.

\begin{table}
\centering
\begin{tabular}{|c|c|c|}
\hline
Model & BLEU & CHRF \\ \hline
Xunzi & 19.67 & 19.36 \\ \hline
Baidu Translate & 23.03 & 22.30 \\ \hline
GPT-4 & 25.12 & 22.64 \\ \hline
AI Taiyan & 36.76 & 31.66 \\ \hline
\end{tabular}
\caption{Results of the Classical-to-Modern translation task.}
\label{table: translation results}
\end{table}

During the automatic evaluation phase, we sampled 100 pairs of Classical Chinese and modern Chinese texts with human-translated references from the Chinese Classics Database. Each segment ranged in length from several dozen to several hundred characters, aiming to evaluate the translation performance of models on both short and long texts. The evaluation metrics used were BLEU and CHRF, which are commonly adopted in the field of machine translation to measure the character-level similarity between machine-generated translations and reference translations; higher values indicate better translation quality. As shown in Table \ref{table: translation results}, ``AI Taiyan'' demonstrates a significant advantage on both metrics.

When analyzing the model output results, we found that although Baidu Translate and GPT-4 received similar scores in automatic evaluations, there are significant differences in their translation strategies: Baidu Translate often exhibits literal copying of the source text, whereas GPT-4 tends to provide more detailed explanations and translations. However, such differences are not captured by BLEU or CHRF metrics. To conduct a more rigorous and accurate evaluation of translation quality, we adopted a human evaluation approach similar to that used for word explanation.

During the manual evaluation phase, the test set similarly comprised texts from two sources: (1) 100 text segments were extracted from several of the most recently published collated editions of ancient Chinese classics, selected in chronological order from the Chinese Classics Database, to form part of the test set; (2) An additional 100 text segments were chosen from extracurricular reading materials and examination-related texts at the secondary school level. In total, the test set consisted of 200 entries. Two examples are provided below.
\begin{itemize}
    \item{晉陵張公治信之明年，皇祐二年也，姦彊帖柔，隱詘發舒，既政大行，民以寧息。夏六月乙亥，大水。公徙囚於高獄，命百隸戒，不共有常誅。夜漏半，水破城，滅府寺，苞民廬居。公趨譙門，坐其下，敕吏士以桴收民，鰥孤老癃與所徙之囚，咸得不死。
    
    (In the second year that Lord Zhang of Jinling governed Xinzhou, which was also the second year of the Huangyou era, those who were previously domineering became compliant, hidden grievances were redressed, and policies of the government were widely implemented. As a result, the people lived in peace and contentment.
    In the sixth month of summer, on the day of Yihai, a great flood occurred. Lord Zhang relocated the prisoners to a high prison and ordered a hundred guards to be vigilant, instructing them not to execute prisoners arbitrarily due to the disaster. In the middle of the night, the floodwaters broke through the city walls, submerging government offices, temples, and the homes of the people. Lord Zhang hurried to the city gate tower and sat beneath it, instructing officials and soldiers to use boats to rescue the people. Thanks to his orders, widowers, orphans, the elderly, the infirm, and the relocated prisoners all survived and did not perish.)}
    \item{順治二年乙酉四月，江都圍急。督相史忠烈公知勢不可為，集諸將而語之曰：“吾誓與城為殉，然倉皇中不可落於敵人之手以死，誰為我臨期成此大節者？”副將軍史德威慨然任之。忠烈喜曰：“吾尚未有子，汝當以同姓為吾後。吾上書太夫人，譜汝諸孫中。”
    
    (In the second year of the Shunzhi reign, in April, Jiangdu was tightly besieged by enemy forces. The chief minister, Lord Shi Zhonglie (Shi Kefa), realized that the situation was hopeless. He gathered the generals and said to them, ``I have sworn to live and die with this city, but in such haste, I cannot allow myself to fall into enemy hands before dying. Who among you is willing to help me accomplish this great act at the critical moment?'' Vice General Shi Dewei readily agreed to take on this responsibility. Lord Shi Zhonglie was pleased and said, ``I have no sons yet. You, sharing the same family name, shall become my heir. I will write to my mother and have you recorded among my descendants in the family genealogy.'')}
\end{itemize}

In the human evaluation experiment, we selected Baidu Translate and GPT-4 as model baselines, both of which demonstrated superior performance in the automatic evaluation. We also invited nine graduate students (master’s and doctoral) majoring in literature and history to provide closed-book answers, which served as the human baseline. The evaluators were four doctoral students majoring in Classical Chinese, Classical Philology, and History. The evaluation methods and procedures were consistent with those used in the word meaning explanation task.

Translation quality was assessed using a five-point scale:
5 points: Very few errors, no critical comprehension errors (including key content words, proper nouns, cohesion, grammatical errors, etc.), semantically fluent and coherent, closely aligned with the original text, and highly supportive of human understanding;
4 points: Few errors, with 1–2 critical errors, semantically fluent and coherent, closely aligned with the original text, and helpful for human understanding;
3 points: Some errors, generally fluent, though with some coherence issues, and moderately supportive of human understanding;
2 points: Numerous errors, lacking fluency, causing reader confusion;
1 point: Extensive errors, fundamentally unintelligible, or completely misleading (harmful nonsense).

Experimental results showed that the overall rating consistency among the four evaluators (measured by Spearman correlation coefficient) reached 0.7548.

\begin{figure}[t]
\centering
\includegraphics[width=0.5\linewidth]{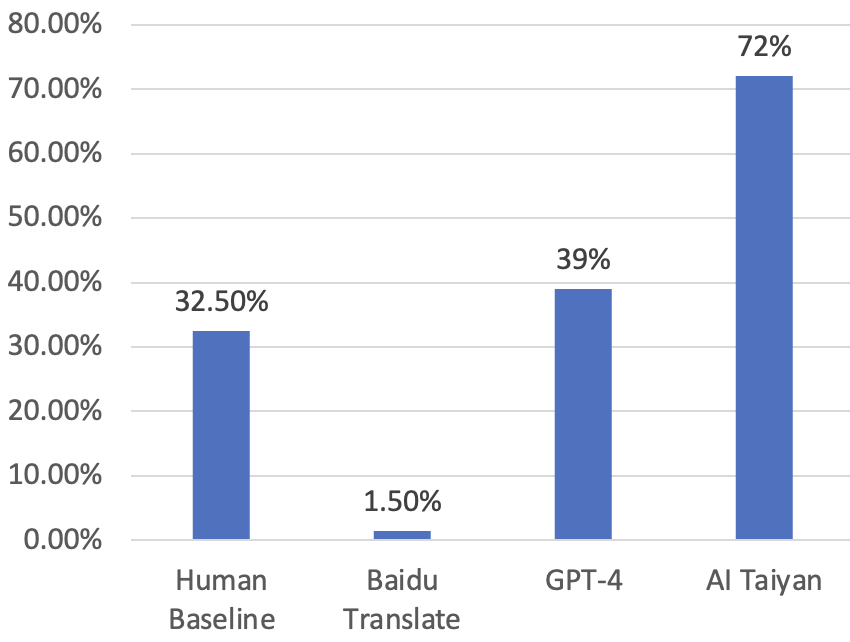} 
\caption{Win rate of models in the classical-to-modern translation task. Mean scores of models: human (3.641), Baidu Translate (1.358), GPT-4 (3.755), AI Taiyan (4.325).}
\label{figure: translation win rate}
\end{figure}

The human evaluation experiment reports results using the commonly adopted ``win rate'' metric in large language model anonymous evaluation \cite{zhao2023survey}. In this context, ``win rate'' refers to the proportion of times each translation method ranks first; if scores are tied, the methods are assigned the same rank. As shown in Figure \ref{figure: translation win rate}, Baidu Translate exhibits the most significant discrepancy between automatic and human evaluation results. Although its automatic evaluation scores are close to those of GPT-4, the actual translation quality is considerably inferior to that of other models. This indicates that relying solely on automatic evaluation metrics for Classical-to-Modern Chinese translation tasks may not yield reliable results. Overall, in the human evaluation stage, ``AI Taiyan'' still demonstrates a clear advantage.
\section{Application}
Based on the evaluation results presented in the previous section, ``AI Taiyan'' demonstrates significant advantages over existing models in various tasks, achieving performance that is close to or even surpasses the human baseline. Given its strong capabilities in analyzing classical texts, this section will further explore its potential applications in fields such as classical text collation, lexicography, and linguistic research.

\subsection{Classical Text Collation}
In the process of collating and publishing ancient texts, it is often necessary for experts to perform tasks such as punctuation, annotation, and translation according to publication requirements. Each of these tasks demands a high level of knowledge and experience from the experts, resulting in substantial human and time costs. ``AI Taiyan'' can play a supportive role at various stages of this workflow, thereby enhancing the efficiency of ancient text collation and publication. In addition, the model can be integrated into digital ancient text application platforms, enabling readers or users to conduct on-demand analysis and obtain personalized annotation content.

In the traditional process of editing and publishing ancient texts, the punctuation stage can benefit from both automated technology and expert review. On one hand, the automatic punctuation technology of ``AI Taiyan'' can be employed for text preprocessing, after which experts can proofread and revise the results to ensure readability and accuracy. On the other hand, this model can also be used during the revision process for post-editing, identifying potential punctuation errors in the manuscript and alerting experts to texts that require special attention.

At the annotation stage, the word explanation function of ``AI Taiyan'' can provide high-quality definitions for terms in the ancient texts. Even when the automatically generated explanations do not fully meet the requirements, editors can still use the Classical-to-Modern Chinese translation results as a reference. In this way, editors can quickly adopt or modify the model suggestions to complete the annotation of key terms.

Generally, edited versions of ancient texts seldom provide full modern translations. This is partly because publishers assume the target audience consists of specialists, so detailed annotations or translations are deemed unnecessary. Another reason is the high difficulty and significant labor and time costs associated with Classical-to-Modern translation. By utilizing the Classical-to-Modern Chinese translation function of ``AI Taiyan'', experts can simply revise the generated translations, significantly reducing the workload. As a result, many edited ancient texts could potentially be transformed into fully translated editions, thereby serving a wider readership.

In the application of digitized ancient texts, different readers encounter varying points of confusion regarding the content. Therefore, providing personalized annotations is particularly important. In this context, the real-time feedback capability of ``AI Taiyan'' becomes especially prominent. It can offer instant modern explanations of characters, words, sentences, and passages according to the reader needs, significantly reducing the difficulty of reading ancient texts and enhancing the overall reading experience.

\subsection{Lexicography}
Lexicographical work involves the interpretation of Classical Chinese vocabulary and the selection of example sentences, tasks that are both highly challenging and labor-intensive. Leveraging ``AI Taiyan'' for large-scale semantic analysis offers significant benefits for both the compilation and revision of dictionaries.

First, given a target keyword, we can collect a large corpus containing this keyword and use ``AI Taiyan'' to annotate the word meaning within its specific contexts. Since context-based semantic interpretation often exhibits considerable diversity, methods such as the Jaro-Winkler distance \cite{winkler1990string} can be employed to cluster these interpretations, with the granularity of clustering adjusted according to specific needs. Consequently, the clustering results can assist experts in determining the frequency and precise usage of each sense, thereby facilitating the arrangement and ordering of dictionary entries.

Second, annotation errors arising from misinterpretations by editors are not uncommon in existing dictionaries, posing significant challenges for dictionary revision. The following two examples illustrate this issue \footnote{From 吳銘, 《漢語大詞典》校札筆記: \href{https://mp.weixin.qq.com/s/3UQNKbqsxXDa70LXAs-Kig}{``進利''}, \href{https://mp.weixin.qq.com/s/7lvANBdQ_gXnBRFJxAhWsQ}{``進資''}.}.
\begin{itemize}
    \item {In the Comprehensive Dictionary of Chinese Words, aka ``Hanyu Da Cidian'', ``進利'' is explained with ``仕進順利'' (promotion goes smoothly). The example sentence is
    
    曄少時，兄晏常云：“此兒進利，終破門戶。”終如晏言。
    
    (When Ye was young, his older brother Yan often said, ``This child is greedy for profit and will eventually ruin the family.'' In the end, things turned out just as Yan had said.)
    
    According to the sentence, ``進利'' means ``貪利'' (greedy for profit).}

    \item{In the dictionary, ``進資'' is explained with ``給予費用'' (provide fund). The example sentence is
    
    爾令行百里，運不絕道，使軍不乏而士益振，以迄有成，賞可後哉！進資一等，以示褒嘉。
    
    (You carried out the order to march a hundred li, with supplies transported continuously, ensuring that the army lacked nothing and the soldiers' morale was further boosted. In the end, you achieved success. Can such a reward be delayed? You are hereby promoted by one rank as a mark of praise and commendation.)
    
    ``進資'' should be explained with ``提升官職'' (promote).}
\end{itemize}

It is evident that the annotation generated by ``AI Taiyan'' can assist editors in clarifying the contextual meaning and avoiding misunderstandings. In the specific process of dictionary revision, this model can be utilized to interpret dictionary entries in conjunction with their example sentences, or to translate the example sentences themselves. Subsequently, the annotated results can be automatically compared with the definitions provided in the dictionary, thereby uncovering inconsistencies and providing valuable clues for the revision work.

\subsection{Linguistic Study}
Compared to humans, a significant advantage of large language models is their ability to process vast amounts of data rapidly. In addition to speed and accuracy in annotation, the use of a single model ensures a high degree of consistency in the results. Taking large-scale word sense annotation as an example, the annotated data can not only support the lexicographical work mentioned above, but also facilitate research into the semantic evolution of Chinese words.

\begin{figure}[t]
\begin{subfigure}
\centering
\includegraphics[width=0.5\linewidth]{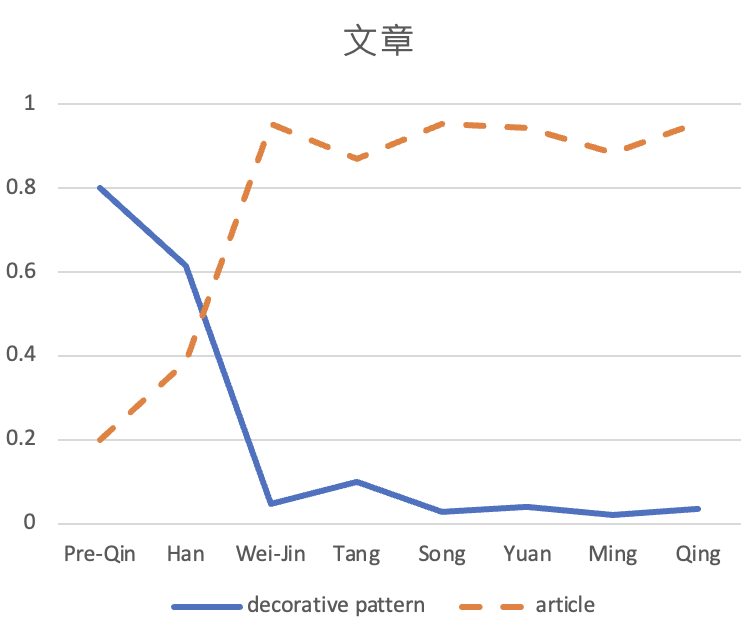} 
\end{subfigure}
\begin{subfigure}
\centering
\includegraphics[width=0.5\linewidth]{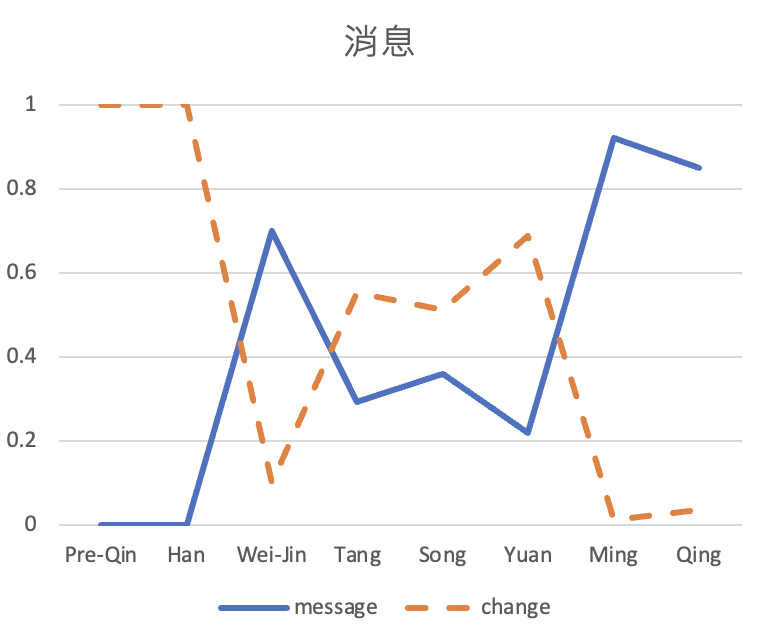} 
\end{subfigure}
\caption{The semantic changes of ``文章'' and ``消息''.}
\label{figure: sense change}
\end{figure}

To conduct relevant research, we constructed a diachronic corpus of Classical Chinese comprising over 100 million characters. The corpus is organized chronologically into the following periods: Pre-Qin, Han, Wei-Jin and Northern and Southern Dynasties, Tang, Song, Yuan, Ming, and Qing. As previously mentioned, given a target keyword, we can retrieve all related corpus data from the database. The model is then required to annotate the senses of the keyword, and subsequently, sense clustering is performed to obtain the distinct senses and their diachronic frequency information. Figure \ref{figure: sense change} provides examples for the words ``文章'' and ``消息'', where the two most frequent senses for each word are selected after being clustered to illustrate their frequency changes over time. According to the statistical results shown, ``文章'' originally referred to decorative patterns and began to be primarily used in literary contexts from the Wei-Jin and Northern and Southern Dynasties onwards. ``消息'' initially referred to the processes of dispersing (``消'') and growing (``息''), frequently used to indicate change. Since the Wei-Jin and Northern and Southern Dynasties, ``消息'' has mainly referred to information, news, and messages. Compared with existing manual and automated analysis methods (e.g., \cite{shu2021}), this approach does not require manually designing sense inventories or annotated data, making it easily extendable to other words. If we apply this method for large-scale automatic annotation and analysis of lexical items, it will undoubtedly facilitate systematic research into the patterns of semantic change in Chinese.
\section{Conclusion}

This study addresses the characteristics of ``low-resource'' and ``knowledge-rich'' scenarios in natural language processing tasks of Classical Chinese, and proposes a method for building a large language model for Classical Chinese from scratch. Firstly, based on the requirements of domain knowledge learning and the current state of available data, a compact large language model architecture is designed (52 layers, 1.8 billion parameters). Subsequently, through data processing, base model training, and finetuning, the ``AI Taiyan'' Classical Chinese large language model is developed. This model demonstrates strong capabilities in interpreting classical texts, supporting a range of challenging Classical Chinese understanding tasks such as sentence segmentation and punctuation, allusion recognition, word explanation, and translation between Classical and modern Chinese. It is compatible with both simplified and traditional Chinese scripts. Experimental results show that, compared with general large models and other domain-specific models, ``AI Taiyan'' exhibits significant advantages across multiple evaluation tasks, achieving performance close to or surpassing human baselines. In addition, this paper discusses the potential applications of it in assisting with classical text collation, dictionary compilation and revision, and linguistic research.

It is worth noting that this study provides a valuable reference for the efficient construction of large language models tailored to specialized domains. Since general-purpose large models often lack a sufficient understanding of domain-specific knowledge when applied to vertical tasks, developing domain-specific LLMs enables a more systematic acquisition of specialized knowledge, thereby better supporting applications in vertical domains. When constructing such specialized models, simply finetuning open-source general models with domain-specific data does not necessarily yield optimal results. In practice, several key issues require particular attention:

First, model developers must collaborate closely with domain experts to clearly define the actual needs of the field and to design relevant tasks. This collaboration ensures that training data is collected and annotated for domain-specific finetuning based on real-world problems. The scale, quality, and diversity of the data have a significant impact on the language capabilities of the model.

Second, it is essential to estimate the required amount of training data and the corresponding model parameter size for different specialized tasks, in order to enhance training efficiency and resource utilization.

Third, after training, the professional capabilities of the model should be evaluated through multiple rounds of testing. Evaluation should not be limited to reporting results on pre-defined test sets, but should also involve domain professionals in human assessments. The evaluations and feedback from domain experts constitute the most valuable source of information for model iteration. Thus, effective cooperation between model developers and domain experts is crucial for the evaluation process.

Fourth, in terms of application, the use of specialized domain LLMs differs from the conversational scenarios typical of general-purpose LLMs. Integrating these models into domain-specific platforms or tools can provide more efficient services for professionals within the field.

It should be noted that the tasks currently handled by large language models remain limited, and these models may still make mistakes on certain issues, potentially leading to misunderstandings. Therefore, at the present stage, the application of such models is primarily positioned as an auxiliary tool. In the future, it will be necessary for large language models for Classical Chinese to incorporate more tasks driven by real-world demands, and to enhance their learning abilities through high-quality data and improved training and fine-tuning mechanisms, so that they can undertake a broader range of tasks related to Classical Chinese.
\end{CJK*}

\begin{CJK*}{UTF8}{gbsn}
\bibliography{scibib}
\bibliographystyle{Science}
\end{CJK*}

\clearpage

\end{document}